\documentclass{article}
\usepackage{ijcai22}
\usepackage[round]{natbib} 
\usepackage[mathscr]{eucal}
\usepackage{epsfig,epsf,psfrag}
\usepackage{amssymb,amsmath,amsthm,amsfonts,latexsym}
\usepackage{amsmath,graphicx,bm,url}
\usepackage[caption=false]{subfig} 
\usepackage{fixltx2e}
\usepackage{array}
\usepackage{verbatim}
\usepackage{bm}
\usepackage{algorithmic}
\usepackage{algorithm}
\usepackage{verbatim}
\usepackage{textcomp}
\usepackage{mathrsfs, bbm,overpic}

\usepackage{stackengine}

\usepackage{tikz}
\usepackage{float}
\usepackage{tabularx}
\usepackage{multirow}
\usetikzlibrary{patterns}

\catcode`~=11 \def\UrlSpecials{\do\~{\kern -.15em\lower .7ex\hbox{~}\kern .04em}} \catcode`~=13 

\allowdisplaybreaks[1]

\newcommand{\bbE}{\mathbb{E}}

\newcommand{\bbR}{\mathbb{R}}

\DeclareMathAlphabet{\mathbsf}{OT1}{cmss}{bx}{n}
\DeclareMathAlphabet{\mathssf}{OT1}{cmss}{m}{sl}

\DeclareSymbolFont{bsfletters}{OT1}{cmss}{bx}{n}  
\DeclareSymbolFont{ssfletters}{OT1}{cmss}{m}{n}
\DeclareMathSymbol{\bsfGamma}{0}{bsfletters}{'000}
\DeclareMathSymbol{\ssfGamma}{0}{ssfletters}{'000}
\DeclareMathSymbol{\bsfDelta}{0}{bsfletters}{'001}
\DeclareMathSymbol{\ssfDelta}{0}{ssfletters}{'001}
\DeclareMathSymbol{\bsfTheta}{0}{bsfletters}{'002}
\DeclareMathSymbol{\ssfTheta}{0}{ssfletters}{'002}
\DeclareMathSymbol{\bsfLambda}{0}{bsfletters}{'003}
\DeclareMathSymbol{\ssfLambda}{0}{ssfletters}{'003}
\DeclareMathSymbol{\bsfXi}{0}{bsfletters}{'004}
\DeclareMathSymbol{\ssfXi}{0}{ssfletters}{'004}
\DeclareMathSymbol{\bsfPi}{0}{bsfletters}{'005}
\DeclareMathSymbol{\ssfPi}{0}{ssfletters}{'005}
\DeclareMathSymbol{\bsfSigma}{0}{bsfletters}{'006}
\DeclareMathSymbol{\ssfSigma}{0}{ssfletters}{'006}
\DeclareMathSymbol{\bsfUpsilon}{0}{bsfletters}{'007}
\DeclareMathSymbol{\ssfUpsilon}{0}{ssfletters}{'007}
\DeclareMathSymbol{\bsfPhi}{0}{bsfletters}{'010}
\DeclareMathSymbol{\ssfPhi}{0}{ssfletters}{'010}
\DeclareMathSymbol{\bsfPsi}{0}{bsfletters}{'011}
\DeclareMathSymbol{\ssfPsi}{0}{ssfletters}{'011}
\DeclareMathSymbol{\bsfOmega}{0}{bsfletters}{'012}
\DeclareMathSymbol{\ssfOmega}{0}{ssfletters}{'012}

\DeclareMathOperator*{\argmin}{arg\,min}

\newcommand{\qednew}{\nobreak \ifvmode \relax \else
      \ifdim\lastskip<1.5em \hskip-\lastskip
      \hskip1.5em plus0em minus0.5em \fi \nobreak
      \vrule height0.75em width0.5em depth0.25em\fi}

\usepackage[colorlinks=true,linkcolor=blue, citecolor=blue, filecolor=magenta, urlcolor=violet]{hyperref}
\usepackage{mathtools}
\mathtoolsset{showonlyrefs}
\newcommand{\Prob}[1]{\mathbb{P}\left( #1 \right)}

\newcommand{\E}{\mathbb{E}}
\newcommand{\R}{\mathbb{R}}
\newtheorem{definition}{Definition}
\newtheorem{lemma}{Lemma}
\newtheorem{proposition}{Proposition}
\newtheorem{theorem}{Theorem}
\newtheorem{corollary}{Corollary}
\newcommand{\cvar}{\mathrm{CVaR}}
\newcommand{\C}{\mathcal{C}}
\newcommand{\cL}{\mathcal{L}}
\newcommand{\vaR}{\textrm{VaR}}
\newcommand{\lcb}{\textrm{LCB}}
\newcommand{\myexp}{\mathsf{e}}
\newcommand{\intinfinity}{\int_0^{\infty}}
\newcommand{\indic}[1]{\mathbb{I}\left\{#1\right\}} 
\usepackage{xspace}
\usepackage{mathtools}

\usepackage{fullpage}
\usepackage[utf8]{inputenc}

\newcommand\blfootnote[1]{%
  \begingroup
  \renewcommand\thefootnote{}\footnote{#1}%
  \addtocounter{footnote}{-1}%
  \endgroup
}

\title{A Survey of Risk-Aware Multi-Armed Bandits}
\author{Vincent Y. F.  Tan$^\dagger$, Prashanth L.A.$^*$,  and Krishna  Jagannathan$^*$ \\ \vspace{.1in} {\normalfont $^\dagger$National University of Singapore, $^*$IIT Madras} \\  \vspace{.05in} {\normalfont \href{mailto:vtan@nus.edu.sg}{vtan@nus.edu.sg}, \href{mailto:prashla@cse.iitm.ac.in}{prashla@cse.iitm.ac.in}, \href{mailto:krishnaj@ee.iitm.ac.in}{krishnaj@ee.iitm.ac.in}}}
\date{\today}

\begin{document}

\maketitle
\begin{abstract}
    
In several applications such as clinical trials and financial portfolio optimization, the expected value (or the average reward) does not satisfactorily capture the merits of a drug or a portfolio.  In such applications, \emph{risk} plays a crucial role, and a risk-aware performance measure is preferable, so as to capture losses in the case of adverse events. This survey aims to consolidate and summarise the existing research on risk measures, specifically in the context of multi-armed bandits.  We review various risk measures of interest, and comment on their properties. Next, we review existing concentration inequalities for various risk measures. Then, we proceed to defining risk-aware bandit problems, We consider algorithms for the regret minimization setting, where the exploration-exploitation trade-off manifests, as well as the best-arm identification setting, which is a pure exploration problem---both in the context of risk-sensitive measures. We conclude by commenting on persisting challenges and fertile areas for future research.\blfootnote{This is an unabridged version of a survey paper of the same title accepted to the 31st International Joint Conference on Artificial Intelligence and the 25th European Conference on Artificial Intelligence (IJCAI-ECAI, 2022).}
\end{abstract}
\section{Introduction}
The multi-armed bandit (MAB) problem studies the problem of online learning with partial feedback that exemplifies the exploration-exploitation tradeoff. This problem has a long history dating back to~\cite{thompson1933likelihood} and finds a wide variety of applications from clinical trials to financial portfolio optimization. In the stochastic MAB problem,  a player chooses or pulls one among several arms, each defined by a certain reward distribution. The player wishes to maximize his reward or find the best arm in the face of the uncertain environment the distributions are {\em a priori} unknown.  There are two general sub-problems in the MAB literature, namely, regret minimization and best-arm identification (also called pure exploration). In the former, in which the exploration-exploitation trade-off manifests, the player wants to maximize his reward over a fixed time period. In the latter, the player simply wants to learn which arm is the best in either the quickest time possible with a given probability of success (the fixed-confidence setting) or he wants to do so with the highest probability of success given a fixed playing horizon (the fixed budget setting). In most of the MAB literature (see \citet{lattimore2020bandit} for an up-to-date survey), the metric of interest is defined simply as the  mean of the reward distribution  associated with the arm pulled.

However, in   real-world applications, the  mean does not satisfactorily capture the merits of   certain actions. In such applications, {\em risk}   plays an important role, and a risk-aware performance measure is often preferred over the average reward. For example, if we have an important face-to-face meeting  downtown, we might want to choose a route that has a slightly higher expected travel time, but whose variance is small. Some  risk measures include the mean-variance \citep{markowitz1952portfolio}, the conditional value-at-risk~\citep{Artzner, rockafellar2000optimization},  spectral risk measures  \citep{acerbi2002spectral}, utility-based shortfall risk~\citep{Artzner,hu2018utility}, and cumulative prospect theory~\citep{tversky1992advances}. This short survey aims to consolidate these various risk measures and to comment on their concentration properties. These properties are then used to describe the latest algorithmic developments for risk-aware bandits in regret minimization  as well as best arm identification settings.

\section{Preliminaries}
For forming   estimators for several risk measures, we require the  empirical distribution function (EDF). 
\begin{definition}
Given  i.i.d.\ samples $\{X_i\}_{i=1}^n$ from the distribution $F$ of a random variable (abbreviated in the following as r.v.) $X$, the EDF is
$$
	F_n (x ) := \frac1{n} \sum_{i=1}^n \indic{X_i \le x} \quad \mbox{for any}\;\;x\in\bbR
$$
\end{definition} 

To establish concentration bounds for  risk measures, one  requires an assumption that bounds the tail of the  distribution. In this article, we consider sub-Gaussian distributions; see Theorem~2.1 of \cite{wainwright2019} for equivalent characterizations. 

\begin{definition}
	A r.v.\ $X$ is {\em sub-Gaussian with (variance proxy) parameter $\sigma^2$} if its cumulant generating function $\log\E [\exp(rX)]\le r^2\sigma^2/2$ for all $r\in\mathbb{R}$. 
\end{definition}
A sub-Gaussian r.v.\ $X$ with parameter $\sigma^2$ satisfies the following bound for any $\epsilon>0$:
\begin{align} 
	\Prob{ X \ge \epsilon} \le \exp\left(-\frac{\epsilon^2}{2\sigma^2}\right). 
	\label{eq:subgauss-1}
\end{align}


For unifying risk measures, we require the notion of the Wasserstein distance, which is defined below.
\begin{definition}
	\label{definition:wasserstein}
	The {\em Wasserstein distance} between two cumulative distribution functions (CDFs) $F_1$ and $F_2$ on $\mathbb{R}$ is
	\begin{align}
		W_{1}(F_1,F_2) := \inf_{F\in\Gamma(F_{1},F_{2})} \int_{\R^2} |x-y|\, \mathrm{d}F(x,y),
	\end{align}
	where $\Gamma(F_1, F_2)$ is the set of couplings of $F_1$ and $F_2$.
\end{definition}
The Wasserstein distance between two CDFs.
 $F_1$ and $F_2$ of two r.v.s $X$ and $Y$, respectively, may be alternatively written as follows:
\begin{align}
&W_{1}(F_1,F_2)=\sup \left|\E(f(X) - \E(f(Y))\right|\nonumber\\*
&=\!\int_{-\infty}^{\infty}|F_{1}(s)\!-\!F_{2}(s)|\mathrm{d}s\!=\!\int_{0}^{1}\!|F_{1}^{-1}(\beta)\!-\!F_{2}^{-1}(\beta)|\mathrm{d}\beta, \nonumber
\end{align}
where the supremum  is over all functions $f:\R\rightarrow\R$ that are $1$-Lipschitz. 
\section{Estimation and concentration of risk measures }

\subsection{Mean-Variance}

We present a measure that effectively balances   between expected reward and variability. Although there are a large number of models that resolve the tension between return and risk---such as the {\em Sharpe ratio} \citep{sharpe1966mutual} and the {\em Knightian uncertainty}~\citep{knight1921risk}, we commence our discussion on one of the most popular  measures, namely the mean-variance   proposed by \cite{markowitz1952portfolio}. 

\begin{definition} \label{def:mv}
    The mean-variance of   a r.v.\ $X$ with mean $\mu$ and variance $\sigma^2$ and risk tolerance $\gamma$ is $\mathrm{MV}=\gamma\mu-\sigma^2$.
\end{definition}


We can recover two extreme cases by considering the extremal values of the risk tolerance $\gamma$. When $\gamma\to\infty$, maximizing $\mathrm{MV}$ simply reduces to maximizing the mean, turning the mean-variance MAB problem into a standard reward maximizing problem. When $\gamma=0$, the mean-variance MAB problem reduces to a variance-minimization problem. When one only has access to samples, one can estimate the mean-variance by using plug-in estimates for the mean and variance. Concentration results for the mean-variance can be found in \cite{zhu2020thompson}. Here, we present a simpler form of the result.

We now provide a concentration result for the mean-variance, assuming that the underlying r.v.\ $X$ is sub-Gaussian. Define the empirical mean-variance $\widehat{\mathrm{MV}}_n := \gamma\hat{\mu}_n - \hat{\sigma}_n^2$ where $\hat{\mu}_n$ and $\hat{\sigma}_n^2$ are the sample mean and sample variance respectively. 
\begin{lemma}
	\label{lemma:subexp-conc}
 Let $X$ be a sub-Gaussian r.v.\ with parameter $\sigma^2$. For any $n\in\mathbb{N}$ and $\epsilon > 0$:
	\begin{align*}
		&\Prob{  |\widehat{\mathrm{MV}}_n -\mathrm{MV} | > \epsilon }   \leq 	 2\exp\left[ - \frac{n\epsilon^2}{8\gamma^2\sigma^2} \right] \\
		&\qquad + 2\exp \left (  - \frac{n}{16} \min \left [ \frac{\epsilon^2}{2\sigma^4}, \frac{\epsilon }{\sigma^2}\right ] \right)   .
	\end{align*} 
\end{lemma}
This result can be proved by using concentration bounds for sub-Gaussian and sub-exponential r.v.s; see, for example    \citet[Prop.~2.2]{wainwright2019}.

\subsection{Lipschitz-continuous risk measures}\label{sec:lipschitz}
In \cite{cassel2018general}, the authors consider general risk measures that satisfy a Lipschitz  requirement under some norm in the space of distributions. 
  \cite{la2020concentration}  use the notion of Wasserstein distance as the underlying norm in defining a continuous class of risk measures as given below.
\begin{definition}\label{def:lip-risk}
	Let $(\cL,W_1)$ denote the metric space of distributions. 
	A risk measure $\rho(\cdot)$ is  Lipschitz-continuous on $(\cL,W)$ if  there exists $L>0$ such that, for any two distributions (CDFs) $F, G \in \cL$,    the following bound holds:
	\begin{align}
		\left| \rho(F) - \rho(G)\right| \le L \ W_1(F,G). \label{tonedef}
	\end{align} 
\end{definition}
Using the EDF $F_n$ computed from $n$ i.i.d.\ samples, we estimate the risk measure $\rho(F)$ satisfying~\eqref{tonedef} as follows: 
\begin{align}
\rho_n := \rho(F_n).
\label{eq:rho-est} 
\end{align}

\begin{theorem}
	\label{thm:subGauss-risk-conc}
	Let $X$ be a sub-Gaussian r.v.\ with parameter $\sigma^2$. Suppose $\rho:\cL\rightarrow \R$ is a Lipschitz risk measure with parameter  $L$. Then, for every $\epsilon$ satisfying $  \frac{512\sigma}{\sqrt{n}}<\frac{\epsilon}{L} < \frac{512\sigma}{\sqrt{n}}+16\sigma\sqrt{\myexp}$, we have
\begin{equation}
	\Prob{|\rho_{n}\!-\!\rho(X)|\!>\!\epsilon}  \leq \exp\bigg(\!- \frac{n}{256\sigma^2\myexp} \Big(\frac{\epsilon}{L}-\frac{512\sigma}{\sqrt{n}}\Big)^2\bigg).\!
\end{equation}
\end{theorem}

Conditional Value-at-Risk \citep{rockafellar2000optimization}, the spectral risk measure \citep{acerbi2002spectral}, and the utility-based shortfall risk \citep{follmer2002convex} are three popular risk measures that are Lipschitz continuous.
 We describe these risk measures below. For other examples of risk measures satisfying \eqref{tonedef}, the reader is referred to  \cite{cassel2018general} and \cite{bhat2019concentration}. 


\subsection{Conditional Value-at-Risk (CVaR)}\label{sec:cvar}
 \begin{definition}
	Let  $(y)^+ := \max(y,0)$. The CVaR at level $\alpha \in (0,1)$ for a r.v.\ $X$ is defined by 
	\[\cvar_{\alpha}(X) := \inf_{\xi\in\bbR} \left\lbrace \xi + \frac{1}{(1-\alpha)}\E\big[\left( X -\xi\right)^+\big] \right\rbrace.\]
\end{definition}

It is well known \citep{rockafellar2000optimization} that the infimum in the definition of CVaR above is achieved for $\xi=\vaR_\alpha(X)$, where $\vaR_\alpha(X)=\inf \lbrace \xi : \Prob{X \leq \xi} \geq \alpha \rbrace$ is the Value-at-Risk (VaR) of  $X$ at confidence level $\alpha$. In financial applications, one prefers CVaR over VaR, since CVaR is coherent, while VaR is not \citep{Artzner}. 
CVaR also admits the following equivalent representation: 
$$ \cvar_{\alpha}(X)  =  \vaR_{\alpha}(X)  + \frac{1}{1 - \alpha} \E \left[ X - \vaR_{\alpha}(X) \right] ^+.$$
The following lemma shows that CVaR is a Lipschitz risk measure in the sense of Definition \ref{def:lip-risk}.
\begin{lemma}
	\label{lemma:cvar-t1}
	Suppose $X$ and $Y$ are r.v.s with CDFs $F_{X}$ and $F_{Y}$, respectively, and $\alpha\in (0,1)$. Then
	\begin{align*}
		|\cvar_{\alpha}(X)-\cvar_{\alpha}(Y)|\leq (1-\alpha)^{-1}W_{1}(F_{X},F_{Y}).
	\end{align*}
\end{lemma}
Let $X_1,\ldots,X_n$ denote i.i.d.\ samples   drawn from the distribution of $X$, and let $\{X_{[k]}\}_{k=1}^n$ denote the order statistics of these samples in decreasing order.   $\cvar_\alpha(X)$ can be estimated as follows:
\begin{equation}
	c_{n,\alpha} = \inf_{\xi \in\mathbb{R}} \bigg\lbrace \xi + \frac{1}{n(1-\alpha)}\sum_{i=1}^n\left( X_i -\xi\right)^+ \bigg\rbrace. \label{cvarest}
\end{equation}
Let  $Y$ denote a r.v.\ with distribution $F_n$, the EDF of $X$. Then, it can be shown that 
\[c_{n,\alpha}\!=\! \cvar_\alpha(Y)\!=\! \hat{v}_{n, \alpha}  + \frac{1}{n( 1\!-\! \alpha)} \sum_{i=1}^n \left( X_i \!-\! \hat{v}_{n,	\alpha} \right) ^+,\]
where $\hat{v}_{n, \alpha}  =  \inf \lbrace x \in\mathbb{R} : F_n(x) \geq \alpha \rbrace$  is the empirical VaR estimate, which can been shown to coincide with the order statistic $X_{[\lfloor n(1-\alpha)\rfloor]}$. 
Thus, the CVaR estimator $c_{n,\alpha}$ coincides with the general template $\rho_n=\rho(F_n)$. 

The result below presents a concentration bound for CVaR estimation under a sub-Gaussianity assumption. 
\begin{corollary}
	\label{cor:cvar-subgauss-wass}
	Suppose that $X$ is a sub-Gaussian r.v.  with parameter $\sigma$. Then, for any $\epsilon$ such that $  \frac{512\sigma}{\sqrt{n}}<(1-\alpha) \epsilon < \frac{512\sigma}{\sqrt{n}}+16\sigma\sqrt{\myexp}$ and $n\ge 1$, we have 
	\begin{align*}
		&\Prob{ \left| c_{n,\alpha} - \cvar_\alpha(X) \right| > \epsilon}\\ &\le \exp\left(- \frac{n}{256\sigma^2\myexp} \left((1-\alpha)\epsilon-\frac{512\sigma}{\sqrt{n}}\right)^2\right). 
	\end{align*}
\end{corollary}
The proof of the bound above follows as a corollary to  Theorem \ref{thm:subGauss-risk-conc} in lieu of Lemma \ref{lemma:cvar-t1}. 

We next present an alternative bound with explicit constants under the assumption that 
the r.v.\ $X$ is continuous with a density $f$ satisfies a growth condition, which is made precise in the lemma below.

\begin{lemma}
	\label{lemma:cvar-subgauss-direct}
	Suppose that the r.v. $X$ is sub-Gaussian with parameter $\sigma$ and continuous, with a density function $f$ that satisfies the following condition: There exist  universal constants $\eta,\delta>0$ such that $f(x)>\eta$ for all $x\in \left[v_\alpha - \frac{\delta}{2}, v_\alpha+\frac{\delta}{2}\right]$. Then, for any $\epsilon > 0$, we have 
\begin{align*}
	&\Prob{ \left| 
		 c_{n,\alpha} - \cvar_\alpha(X) \right| > \epsilon} 
		 \le 2\exp\left[-\dfrac{n\epsilon^2(1-\alpha)^2}{8\sigma^2}\right]	 \\
		 &\qquad\qquad + 4 \exp \left[  -\frac{n (1-\alpha)^2\eta^2\min\left(\epsilon^2,4\delta^2\right)}{64}   \right].
	\stepcounter{equation}\tag{\theequation}\label{eq:cvar-subgauss-conc}
\end{align*} 
\end{lemma}
The reader is referred to \cite{prashanth2020concentration} for a proof of the result above.
Other works on CVaR estimation can be found in  \cite{brown2007large,wang2010deviation,kolla2019concentration,thomas2019concentration,kagrecha2019distribution} and \cite{prashanth2020concentration}.


\subsection{Spectral risk measure (SRM)}
Given a risk spectrum $\phi:[0,1]\rightarrow [0,\infty)$, the SRM $M_{\phi}$ associated with $\phi$ is defined  by  \citep{acerbi2002spectral}
\begin{align}
	M_{\phi}(X) := \int_{0}^{1}\phi(\beta)F_X^{-1}(\beta)\,\mathrm{d}\beta. \label{specdef}
\end{align}
$M_{\phi}$ is a coherent risk measure if the  risk spectrum is increasing and integrates to 1. Further, $M_{\phi}$ generalizes CVaR, since setting $\phi(\beta)=(1-\alpha)^{-1}\indic{\beta\geq \alpha}$ leads to $M_{\phi}=\cvar_{\alpha}(X)$.

The result below shows that an SRM is a Lipschitz risk measure if the risk spectrum $\phi$ is bounded.
\begin{lemma}
	\label{lemma:srm-t1}
	Suppose $\phi(u)\leq K$ for all $u\in[0,1]$, and let $X$ and $Y$ be r.v.s with CDFs $F_{X}$ and $F_{Y}$, respectively.  Then 
	\begin{equation}
		|M_{\phi}(X)-M_{\phi}(Y)|\leq K\ W_{1}(F_{X},F_{Y}).
		\label{specrsklip}
	\end{equation}
\end{lemma}

Specializing the estimator $\rho_n = \rho(F_n)$ for SRM leads to the following estimator:
\begin{equation}
	m_{n,\phi}=\int_{0}^{1}\phi(\beta)F_{n}^{-1}(\beta)\,\mathrm{d}\beta. \label{empspecdef}
\end{equation}
Using Lemma \ref{lemma:srm-t1} and Theorem \ref{thm:subGauss-risk-conc}, it is straightforward to obtain the following concentration bound for SRM estimation.
\begin{corollary}
	\label{cor:srm-subgauss-wass}
	Suppose that $X$ is a sub-Gaussian r.v.  with parameter $\sigma$. Then, for any $\epsilon$ such that $  \frac{512\sigma}{\sqrt{n}}<\frac{\epsilon}{K} < \frac{512\sigma}{\sqrt{n}}+16\sigma\sqrt{\myexp}$ and $n\ge 1$, we have 
	\begin{align*}
		&\Prob{ \left| m_{n,\phi} \!-\! M_{\phi}(X) \right| > \epsilon}  \\
		&\le \exp\left(- \frac{n}{256\sigma^2\myexp} \left(\frac{\epsilon}{K}-\frac{512\sigma}{\sqrt{n}}\right)^2\right).
	\end{align*}
\end{corollary}

Since CVaR and spectral risk measures are weighted averages of the underlying distribution quantiles, a natural alternative to a Wasserstein distance-based approach is to employ concentration results for quantiles;
cf.~\cite{prashanth2020concentration,pandey2021estimation}. While such an approach can provide bounds with better constants, the resulting bounds also involve distribution-dependent quantities. In this survey, we have chosen the Wasserstein distance-based approach since it provides a unified bound for an abstract risk measure that is Lipschitz, and one can easily specialize this bound to handle CVaR, SRM and other risk measures. Secondly, a template confidence-bound type bandit algorithm can be arrived at using the unified concentration bound in Theorem \ref{thm:subGauss-risk-conc}.  
 \subsection{Utility-based shortfall risk (UBSR)}
 In financial applications, one would be interested in bounding a high loss that occurs with a given low probability. This amounts to bounding the VaR of the loss distribution, at a pre-specified level $\alpha$. VaR as a risk measure is not preferable as it violates sub-additivity, which implies diversification could increase risk. In \cite{Artzner}, the authors identified certain desirable properties for a risk measure, namely,  monotonicity, sub-additivity, homogeneity, and translational invariance. A risk measure satisfyingt these properties is said to be \textit{coherent}. CVaR is a risk measure that is closely related to VaR, and is also coherent.  
 Convex risk measures  \cite{follmer2002convex} are a further generalization of risk measures beyond coherence, because sub-additivity and homogeneity imply convexity. A popular convex risk measure is UBSR. We introduce this risk measure below.

 For a r.v.\ $X$, the UBSR $S_\alpha(X)$ is defined as follows:
 \begin{align}
 	S_\alpha(X) := \inf\left\{\xi \in \R : \E\left[  l(X-\xi)\right] \le \alpha\right\},
 	\label{eq:ubsr-def}
 \end{align}
 where $l:\R\rightarrow \R$ is a utility function. For the purpose of showing that UBSR is a Lipschitz risk measure, we require that $l$ is Lipschitz. While our results below require no additional assumptions on $l$, we point out that a convexity assumption on $l$ leads to a useful dual representation  \citep{follmer2002convex}. Further, if the utility is increasing, then the estimation of  UBSR from i.i.d.\ samples can be performed in a computationally efficient manner; see \cite{hu2018utility}.


 The following lemma shows that UBSR is a Lipschitz risk measure in the sense of Definition \ref{def:lip-risk}. 
 \begin{lemma}
  	\label{lemma:ubsr-t1}
  	Suppose the utility function $l$ is non-decreasing, and there exist $K,k>0$ such that $l$ is $K$-Lipschitz and satisfies $l(x_{2})\geq l(x_{1})+k(x_{2}-x_{1})$ for every $x_{1},x_{2}\in\R$ satisfying  $x_{2}\geq x_{1}$. Let $X$ and $Y$ be  r.v.s with CDFs $F_{X}$ and $F_{Y}$, respectively. Then
  	\begin{align}
  		|S_{\alpha}(X)-S_{\alpha}(Y)|\leq \frac{K}{k} W_{1}(F_{X},F_{Y}).
  		\label{ubrlip}
  	\end{align}
 \end{lemma}
 To estimate the UBSR $S_\alpha(X)$ from  $n$ i.i.d.\ samples $X_1, \ldots, X_n$, we use the following sample-average approximation to the optimization problem in~\eqref{eq:ubsr-def} (also see \cite{hu2018utility}):
 \begin{align}
 	\min_{\xi\in \R} \; \xi  \quad\textrm{ subject to }\quad \frac{1}{n}\sum_{i=1}^n  l(X_i-\xi)\le \alpha.
 	\label{eq:sr-est}
 \end{align}
 As in the case of CVaR and SRM, the UBSR estimator provided above is a special case of the general estimate given  by $\rho_n := \rho(F_n)$.
 Further, the estimation problem in \eqref{eq:sr-est} can be solved in a computationally efficient fashion using a bisection method; see \cite{hu2018utility}.

 Using Lemma \ref{lemma:ubsr-t1} and Theorem \ref{thm:subGauss-risk-conc}, we obtain the following concentration bound.
 \begin{corollary}
 	\label{cor:ubsr-subgauss-wass}
 	Suppose that $X$ is a sub-Gaussian r.v.\ $X$  with parameter $\sigma$. Then, for any $\epsilon$ such that $  \frac{512\sigma}{\sqrt{n}}<\frac{\epsilon k}{K} < \frac{512\sigma}{\sqrt{n}}+16\sigma\sqrt{\myexp}$ and $n\ge 1$, we have 
 	\begin{align*}
 		&\Prob{ \left| \xi_{n,\alpha}  - S_\alpha(X) \right| > \epsilon}\\
 		& \le \exp\left(- \frac{n}{256\sigma^2\myexp} \left(\frac{k\epsilon}{K}-\frac{512\sigma}{\sqrt{n}}\right)^2\right). 
 	\end{align*}
 \end{corollary}

 Next, we present an example of a risk measure that is {\em not} Lipschitz.
 	Consider a risk measure $\rho(F) = \int_0^\infty w(1-F(x))\, \mathrm{d}x$, where $w$ is a weight function. Letting $w(p)= \sqrt{p}$, we claim there does not exist a norm $\| \cdot \|$, on the space of distribution functions, such that the following condition holds for all distributions functions $F_1,F_2$ of non-negative r.v.s:
 		\begin{align}
 			\left| \rho(F_1) - \rho(F_2)  \right| \le  L \| F_1- F_2 \|.\label{eq:ss}
 		\end{align}
 	Consider two Bernoulli distributions $F_1$ and $F_2$, with parameters $p_1$ and $p_2$, respectively. It is easy to see that 
 	\[ \rho(F_1) = w(p_1) \; \textrm{ and }\; \rho(F_2) = w(p_2).\]	
 	Setting $p_1=4\epsilon$ and $p_2=\epsilon$, for some $\epsilon > 0$, we claim that the condition in \eqref{eq:ss} does not hold for $F_1$ and $F_2$. To see this, observe that the left-hand side of \eqref{eq:ss} is $\sqrt{4\epsilon} - \sqrt{\epsilon} = \sqrt{\epsilon}$.  On the other hand,  $(F - G)(x)  = \epsilon \mathbbm{1}_{[0,1)}(x)$. By the homogeneity property of norms, $\|F - G\|$ must scale linearly with $\epsilon$, and hence the inequality in \eqref{eq:ss} cannot hold for sufficiently small~$\epsilon$. Thus this risk measure $\rho(\cdot)$  violates the Lipschitz condition in Definition \ref{def:lip-risk}.

While the above measure appears contrived, it is not the case. The $\rho(\cdot)$ defined above is closely related to a risk measure based on cumulative prospect theory, where one employs a weight function to distort cumulative probabilities. We describe such a risk measure in the next section.


\subsection{Risk measures based on cumulative prospect theory (CPT)}
 We present a risk measure based on CPT, and this risk measure does not satisfy the Lipschitz condition in Definition \ref{def:lip-risk}.  The CPT-value of an r.v.\ $X$ is defined as \citep{prashanth2015cumulative}
	\begin{align}
		\C(X) &:= \intinfinity w^+\left(\Prob{u^+(X)>z}\right) \,\mathrm{d}z \nonumber\\
		&\qquad - \intinfinity w^-\left(\Prob{u^-(X)>z}\right) \,\mathrm{d}z, \label{eq:cpt-general}
\end{align}
where
$u^+,u^-:\R\rightarrow \R_+$ are the utility functions that are assumed to be continuous,
with $u^+(x)=0$ when $x\le 0$ and increasing otherwise, and with $u^-(x)=0$ when $x\ge 0$ and decreasing otherwise,
$w^+,w^-:[0,1] \rightarrow [0,1]$ are weight functions assumed to be continuous, non-decreasing and satisfy $w^+(0)=w^-(0)=0$ and $w^+(1)=w^-(1)=1$.

Given a set of i.i.d.\ samples $\{X_i\}_{i=1}^{n}$, we form the EDFs $F_n^+(x)$ and $F_n^-(x)$  of the r.v.s $u^+(X)$ and $u^-(X)$, respectively.
Using these EDFs, we estimate CPT-value as follows:
\begin{align}
	\overline \C_n \!=\! \intinfinity \!\!\!\! w^+\left(1\!-\!F_n^+\left(x\right)\right)  \,\mathrm{d}x  \!-\! \intinfinity\!\!\!\! w^-\left(1\!-\! F_n^-\left(x\right)\right)   \mathrm{d}x, \label{eq:cpt-est}
	\end{align}
The integrals  above can be computed using the order statistics; see \cite{prashanth2018cpt}. 
 More precisely, let $X_{[1]}\le X_{[2]} \le \ldots \le X_{[m]}$ denote the order statistics. Then, the first and second integrals in \eqref{eq:cpt-est}, denoted by $\overline \C_m^+$ and $\overline \C_m^-$, respectively, are estimated as follows:
 	\begin{align*}
 		\overline \C_n^+&:=\sum_{i=1}^{n} u^+(X_{[i]}) \bigg(w^+\!\Big(\frac{n+1-i}{n}\Big)\!-\! w^+\!\Big(\frac{n-i}{n}\Big) \bigg),\\
 		\overline \C_n^-&:=\sum_{i=1}^{n} u^-(X_{[i]}) \bigg(w^-\Big(\frac{i}{n}\Big)- w^-\Big(\frac{i-1}{n}\Big) \bigg). 
 \end{align*}

We now present a concentration result for CPT-value estimation assuming bounded support for the underlying distribution; see \cite{prashanth2015cumulative,prashanth2018cpt} for the proofs. We start by stating some assumptions.

\noindent\textbf{(A1)} The weight functions $w^{\pm}$ are H\"{o}lder continuous with common order $\alpha$ and constant $H$, i.e., 
$$\sup_{x,y \in [0,1]:x\ne y} \frac{| w^{\pm}(x) - w^{\pm}(y) |}{| x-y |^{\alpha}} \leq H;$$ 

\noindent\textbf{(A2)} The utility functions $u^+,u^-:\R\rightarrow \R_+$ are bounded above by $M$.

\begin{proposition}
	\label{prop:cpt-est}
	Under (A1) and (A2), 
 for all $ \epsilon >0$, 
	\begin{align}
		\Prob{\left| \overline \C_n \!-\! \C(X) \right|\!\geq\!  \epsilon}\le
		2\exp\bigg(\! -2n\Big(\frac{\epsilon}{HM}\Big)^{\frac{2}{\alpha}}\bigg).
		\label{eq:cpt-conc-bounded}
	\end{align}	
\end{proposition}
 	The proof is based on the Dvoretzky--Kiefer--Wolfowitz (DKW) inequality, which establishes concentration of the EDF around the true  distribution.
For the sub-Gaussian case, one can use a truncated CPT-value estimator, and establish an exponentially decaying tail bound. 
In this case, we use a truncated CPT-value estimator \citep{bhat2019concentration}. 
 For a given truncation level $\tau_n$ (to be specified later),  let $\left(F_n|_{\tau_n}\right)^+$ and $\left( F_n|_{\tau_n}\right)^-$ denote the truncated EDFs formed from the samples $\{u^+(X_i)\}_{i=1}^n$ and $\{u^-(X_i)\}_{i=1}^n$, respectively.
 Using the truncated EDFs, the CPT-value is estimated as follows:
 \begin{align}
 	C_n =& \int_{0}^{\tau_n} w^+(1-\left(F_n|_{\tau_n}\right)^+(x))  \,\mathrm{d}x\nonumber\\
 	&\quad- \int_{0}^{\tau_n} w^-(1-\left(F_n|_{\tau_n}\right)^-(x))  \,\mathrm{d}x.\label{eq:cpt-est=trunc}
 \end{align}
 In comparison to \eqref{eq:cpt-est}, we have used complementary (truncated) EDFs in the estimator defined above.

 \noindent\textbf{(A2')} The utility functions $u^+,u^-:\R\rightarrow \R_+$ are sub-Gaussian with parameter $\sigma^2$.

 \begin{proposition}
 	\label{cor:cpt-subexp}
 	Assume (A1) and (A2'). Set $\tau_n = \left(\sigma\sqrt{\log n} + 1\right)$. For all $ n\geq 1$  form the CPT-value estimate $C_n$ using \eqref{eq:cpt-est}. Then,  we have the following bound for $\epsilon > \frac{8H}{\alpha n^{\alpha/2}}$:
 	\begin{align*}
 		&\Prob{\left|{C}_n - C(X)\right|> \epsilon}
 		  \\
 		&\le 2\exp\left( -2 n \left(\frac{2}{H^2\log n}\right)^{\frac1{\alpha}} \left(\epsilon - \frac{8H}{\alpha n^{\alpha/2}} \right)^{\frac{2}{\alpha}}  \right).
 	\end{align*}
 \end{proposition}


\section{Regret minimization}
We now discuss how the above-mentioned concentration bounds are used in the context of   regret minimization in the theory of MABs. 
A {\em bandit instance} $\nu$ is a collection of $K$ distributions $(\nu_1, \nu_2, \ldots, \nu_K)$. Each distribution $\nu_i$ has mean $\mu_i \in \mathbb{R}$ and without loss of generality, one may assume that $\mu_1\ge\mu_2\ge \ldots\ge\mu_K$. The agent interacts with the bandit environment at each time  by pulling an arm $A_t \in \{1,\ldots, K\}$ and observing a reward $X_{t,A_t}$ drawn from $\nu_{A_t}$. The decision of which arm to pull depends on the previous arm pulls as well as rewards $\mathcal{H}_{t-1} = ( A_1, X_{1,A_1}, \ldots A_{t-1}, X_{t-1, A_{t-1}}  ) $. Over a fixed time period (or horizon) of length $n$, the agent desires to maximize his/her total reward $\sum_{t=1}^n X_{t,A_t}$ by designing an appropriate policy $\pi = \{\pi_{t}\}_{t=1}^n$ where $\pi_t$ is $\sigma(\mathcal{H}_{t-1}) $-measurable. An equivalent measure is the {\em regret}
$$
R_n(\nu,\pi ): = \bbE\bigg[n\mu_1 - \sum_{t=1}^n \mu_{A_t}\bigg],
$$
which is to be minimized over $\pi$. 
However, this measure pays no attention to {\em risk}, which is the central theme of this survey. More generally, the agent would like to minimize the $\rho$-regret 
$$
R_n^\rho(\nu,\pi ):= \bbE\bigg[ n \max_{1\le i \le K}\rho(\nu_i) - \sum_{t=1}^n \rho(\nu_{A_t})  \bigg],
$$
where $\rho(\nu)$ is an appropriate risk measure. Let $\Delta_i = \rho(\nu_{i^*})-\rho(\nu_i)$ be the gap  between the risk of the best arm $i^*$ and that of the $i$-th arm.   



\subsection{Confidence bound-based algorithms} 
We present an adaptation Risk-LCB of the well-known UCB algorithm~\citep{auer2002finite} to handle an objective based on   abstract risk measures $\rho$. The algorithm caters to Lipschitz risk measures (see Definition~\ref{def:lip-risk}) and arms' distributions that are sub-Gaussian. Here, note that we are {\em minimizing} the risk measure instead of {\em maximizing} the reward.

To obtain the confidence widths, we first rewrite the concentration bound in Theorem \ref{thm:subGauss-risk-conc} as follows: For any $\delta \in [\exp(-n),1]$, with probability  $\ge1-\delta$,
\[	|\rho_{m}-\rho(X)|\le  \frac{L\sigma\left(\sqrt{ 256\myexp\log(\frac{1}{\delta})} +512\right)}{\sqrt{m}},\]
where $\rho_m=\rho(F_m)$ is an $m$-sample estimate of the risk measure $\rho(X)$. The range of $\delta$ is restricted to $[\exp(-n),1]$ due to the following constraint on $\epsilon$ in Theorem \ref{thm:subGauss-risk-conc}: 
\begin{align}  
\frac{512\sigma}{\sqrt{n}}<\frac{\epsilon}{L} < \frac{512\sigma}{\sqrt{n}}+16\sigma\sqrt{\myexp}.
\label{eq:eps-constraint}
\end{align}

For the classic bandit setup with a expected value objective, the finite sample analysis provided by \cite{auer2002finite} chose to set $\delta =   t^{-4}$ for obtaining a sub-linear regret guarantee.
In our setting, we set $\delta=\frac{8}{t^4}$ as this choice satisfies the constraint coming from \eqref{eq:eps-constraint}, in turn facilitating the application of the high confidence guarantee on $\rho_m$ given above. Under this choice of $\delta$, in any round $t$ of Risk-LCB and for any arm $k \in \{1,\ldots,K\}$, we have 
\begin{align}  
	&	 \Prob{ \rho(i) \in [ \rho_{i,T_{i}(t-1)} \pm w_{i,T_{i}(t-1)} ] 
	} \ge 1- 8 t^{-4}, \textrm{ where} \nonumber\\
	& 	\qquad\quad	w_{i,T }  :=  \frac{L\sigma \left(32\sqrt{\myexp\log t} + 512\right)}{\sqrt{T} }.\label{eq:lcb-w}
\end{align}	 		
In the above,
$\rho_{i,T_{i}(t-1)}$ is the estimate of the risk measure for arm $i$ computed using $\rho_n = \rho(F_n)$ from $T_i(t-1)$ samples.

The Risk-LCB algorithm would play all arms once in the initialization phase, and in rounds $t\ge K+1$, play an arm $A_t$ according to the following rule:
\[A_t = \argmin\limits_{1\le i\le K} \left(\lcb_t(i) := \rho_{i,T_{i}(t-1)} - w_{i,T_{i}(t-1)}\right).\]
\begin{theorem}
	\label{thm:gap-cvar-regret}
	Consider a $K$-armed stochastic bandit problem with a Lipschitz risk measure $\rho$ and the arms' distributions are sub-Gaussian  with parameter~$\sigma^2$. Then, the expected regret $R_n^\rho$ of Risk-LCB satisfies the following bound:
	\[ R_n^\rho \! \le \! \sum_{i:\Delta_i>0 } \left(\dfrac{4\sigma^2 L^2 [32\sqrt{\myexp\log n}\!+ \! 512]^2}{\Delta_i}  + 28K\Delta_i\right).\]
		Further, $R_n$ satisfies the following bound that does not scale inversely with the gaps:
	 	\begin{align*} 
	 	&\E (R_n) \\ 
	 	&\le \left(2  \sigma L (32\sqrt{\myexp\log n}+512) 
	 	+ 6  \sum_{i:\Delta_i>0 }\Delta_i\right) \sqrt{K n}.
	 	\end{align*}
\end{theorem}  
This bound mimics that of risk-neutral UCB except that the $\Delta_i$'s depend on $\rho$.

The regret bound derived in Theorem~\ref{thm:gap-cvar-regret} is not applicable for CPT, as it is not a Lipschitz risk measure. However, one can derive a confidence bound-based algorithm with CPT as the risk measure using the result in Proposition \ref{prop:cpt-est} for the case of arms' distributions with bounded support, see \cite{aditya2016weighted}. 
The regret bound is of the order $O\big(\sum_{i:\Delta_i>0 }\frac{\log n}{\Delta_i^{{2}/{\alpha}-1}} \big)$, where $\alpha$ is the  H\"{o}lder exponent (see {\bf (A1)} above). This bound cannot be improved in terms of dependence on the gaps and horizon $n$; see Theorem 3 in \cite{aditya2016weighted} for details.
For an extension of confidence  bound-based algorithms to handle sub-Gaussian arms' distributions, see \cite{la2020concentration}.

\subsection{Thompson sampling}
Another popular class of algorithms for regret minimization  is   {\em Thompson sampling}~\citep{thompson1933likelihood,agrawal2012analysis,kaufmann2012thompson,russo2018}. Here, one starts with  a prior over the set of bandit environments. In each round,  as observations are obtained, the prior is updated to a posterior. The agent then samples an environment from the posterior and chooses the action that is optimal for the sampled environment.

The first attempt at using Thompson sampling for risk-averse bandits  was by \cite{zhu2020thompson} in which the authors applied    Thompson sampling  to mean-variance bandits for Gaussian and Bernoulli arms. This work is the Thompson sampling analogue of the UCB-based ones for the mean-variance by  \cite{sani2013risk} and \cite{valiki15}. For  the Gaussian case, \cite{zhu2020thompson} considered sampling the precision (inverse variance) of the Gaussian $\tau_{i,t}$ of arm $i$ at time $t$ from a Gamma distribution with parameters $\alpha_{i,t}$ and $\beta_{i,t}$.  Subsequently, the mean of the Gaussian $\theta_{i,t}$  is sampled from a Gaussian whose mean is set to be the empirical mean and whose variance is the inverse of the number of times arm $i$ has been played. The Gamma and Gaussian are  chosen as they are conjugate to the precision and mean of the Gaussian respectively.  This algorithm, termed MVTS (for mean-variance Thompson sampling), has expected regret satisfying
\begin{align}
    \!\limsup_{n\to\infty}\frac{R_n^\rho }{\log n}
    \!\le\!\sum_{i=2}^K\! \max\bigg\{ \frac{2}{\Gamma_{1,i}^2} , \frac{1}{h(\frac{\sigma_i^2}{\sigma_1^2})} \bigg\} \big(\Delta_i \!+\! 2\overline{\Gamma}_{i}^2 \big), \!\label{eqn:zhu_tan}
\end{align}
where $\Gamma_{i,j}:=\mu_i-\mu_j$ is the gap between the means of arms $i$ and $j$, $\overline{\Gamma}_{i}^2:= \max_{ j = 1,\ldots,K} (\mu_i-\mu_i)^2$, $\Delta_i := \mathrm{MV}_{i^*}-\mathrm{MV}_i$ is the gap between the mean-variance  of arm  $i$ and the arm with the best MV $i^*$, and $h(x) := \frac{1}{2} (x-1-\log x)$. This bound was shown to be order-optimal in the sense of meeting a problem-dependent information-theoretic lower bound as~$\gamma$  tends to either $0$ or diverges to~$\infty$.

\cite{baudry2020thompson} considered an MAB problem in which the quality of an arm is measured by the CVaR  of the reward distribution. They proposed B-CVTS and M-CVTS for continuous bounded   and multinominal rewards, respectively. Leveraging novel  analysis by~\cite{pmlr-v117-riou20a}, the authors bounded the regret performances of both algorithms and show that they are asymptotically optimal. This is    desirable  as it improves on \cite{zhu2020thompson}, albeit for different risk measures, since the level $\alpha$ does not have to tend to its extremal values to be asymptotically optimal. 
Generalizations of and guarantees for Thompson sampling algorithms    for other risk measures,  such as the empirical distribution performance measures (EDPMs) and distorted risk functionals (DRFs) proposed in \cite{cassel2018general} and \cite{huang2021offpolicy} respectively are discussed in \citet{chang2022}.

\section{Pure exploration} \label{sec:bai}

We  now review existing work on risk-aware \emph{pure exploration} in MABs. Risk-aware pure exploration problems typically involve finding the best arm (or the best few arms), where the merit of an arm is defined to accommodate a risk measure or a constraint. Pure exploration problems have been studied in two settings, namely, \emph{fixed budget} and \emph{fixed confidence}.

\subsection{Fixed budget algorithms}
In the fixed budget setting, the objective is to find the arm (or the best few arms) using a fixed number of arm plays, so as to minimize the probability of misidentification. To be more precise, let $n$ be the fixed ``budget'', i.e., the total number of arm plays allowed. In the notation of the previous section, let $\rho(\nu_k)$ be the risk measure of interest associated with arm $k$, and let $k^*$ be the optimal arm (often assumed to be unique for simplicity). A given arm selection policy $\pi = \{\pi_{t}\}_{t=1}^n$ samples the arms as before, and at the end of the budget, identifies the arm $k_n^\pi$ to be the optimal arm. The merit of the arm selection algorithm is captured by the probability of misidentifying the best arm, which is simply $\Prob{k_n^\pi\neq k^*}.$

The recent paper by \cite{zhang2021quantile} considers pure exploration in a fixed budget setting, where the risk measure is the VaR or quantile at a particular level. Specifically, given a quantile level $\alpha,$ the objective is to identify a set of $m$ arms with the highest VaR$_\alpha$ values within a fixed budget $n$. The authors obtain the empirical VaR estimate using the order statistic $X_{(\lfloor n(1-\alpha)\rfloor)}$ as explained in Section~\ref{sec:cvar}. They then derive exponential concentration bounds for the VaR$_\alpha$ estimate under the assumptions that the underlying r.v.s have an increasing hazard rate, and a continuously differential probability density function. A similar VaR concentration result is derived in Prop.~2 of \cite{kolla2019concentration} under milder assumptions---nevertheless, VaR concentration results necessarily involve constants that depend on the slope of the distribution in the neighbourhood of the $\alpha$-quantile.

The algorithm proposed in \cite{zhang2021quantile} called Q-SAR (Quantile Successive Accept-Reject) adopts the well-known algorithm of \cite{bubeck2013multiple} to the risk-aware setting.   Q-SAR  first divides the time budget $n$ into $M-1$ phases, where the number of samples drawn for each arm in each phase is chosen exactly as in \cite{bubeck2013multiple} for the risk neutral setting. At each
phase $p\in\{1,2,\dots,M-1 \},$ the algorithm maintains two sets of arms, namely the \emph{active set}, which contains all arms that are actively drawn in phase $p,$ and the \emph{accepted set} which contains arms that have been accepted. During each phase, based on the empirical VaR estimates, an arm is removed from the active set, and it is either accepted or rejected. If an arm is accepted, it is added into the accepted set. When the time budget ends, only one arm remains in the active set, which together with the accepted set, forms the recommended set of best arms. In Theorem~4 of \cite{zhang2021quantile} the Q-SAR algorithm is shown to have an exponentially decaying probability of error in the budget $n,$ using the VaR concentration bound. 

Other   papers on risk-aware pure exploration with fixed budget include \cite{kagrecha2019distribution} and \cite{prashanth2020concentration}, which consider CVaR as the risk measure. In \cite{prashanth2020concentration}, the authors derive concentration bounds for CVaR estimates, considering separately sub-Gaussian, light-tailed (or sub-exponential) and heavy-tailed distributions.  For the sub-Gaussian and light-tailed cases, the estimates are constructed as described in Section~\ref{sec:cvar}. For heavy-tailed random variables, they assume a bounded moment condition, and derive a concentration bound for a truncation-based estimator. For all three distribution classes, the concentration bounds exhibit exponential decay in the number of samples. The authors then consider the best CVaR-arm identification problem under a fixed budget. The algorithm, called CVaR-SR, adopts the well-known successive rejects algorithm from \cite{audibert2010best} to best CVaR arm identification. Exponentially decaying bounds on the probability of incorrect arm identification are derived using the CVaR concentration results.  We remark that the Wasserstein distance approach outlined in Section~\ref{sec:lipschitz} gives a direct CVaR concentration bound for sub-Gaussian distributions, but does not seem to extend easily for distributions with heavier tails.

In \cite{kagrecha2019distribution}, the authors consider a risk-aware best-arm identification problem, where the merit of an arm is defined as a linear combination of the expected reward and the associated CVaR. A notable feature in this paper is the \emph{distribution obliviousness} of the algorithms, i.e., the algorithm (which is again a variant of successive rejects) is not aware of any information on the reward distributions, including tails or moment bounds.

We conclude this subsection by commenting on a common desirable feature in all the algorithms reviewed above. The SR (and SAR) family of algorithms does not require knowledge of the constants in the concentration bounds for their implementation. These constants, which are often distribution dependent, are not known in practice. They appear only in the analysis and the upper bound on the probability of the error. This is in contrast with confidence intervals-based algorithms used for regret minimization, where the distribution dependent constants from the concentration bound do appear in the confidence term, necessitating the knowledge of these constants in the algorithm's implementation.  
\subsection{Fixed confidence algorithms}

In fixed confidence pure exploration, the objective is to identify with a high degree of confidence (say with probability at least $1-\delta$) the best arm with the smallest possible number of arm plays. A related (and less stringent) objective is called PAC (Probably Approximately Correct), where the objective is to identify as quickly as possible, the set of arms with reward values that are $\epsilon$-close to the best arm, with a confidence level at least $1-\delta.$ More precisely, a policy in the fixed confidence setting consists of a \emph{stopping time} $T$ (defined w.r.t.\ the filtration $\sigma(\mathcal{H}_{t-1})$ from the previous section), and an arm selection rule  $\{\pi_{t}\}_{t=1}^T.$ The objective is to ensure that the arm identified at the stopping time $T$ is the `true best arm' $k^*$ with probability at least $1-\delta.$ The figure of merit is the expected sample complexity $\mathbb E [T],$ which should be as small as possible for a fixed $\delta.$

In~\cite{szorenyi2015qualitative}, the authors consider PAC best arm identification, where the quality of an arm is defined in terms of  VaR (or quantile) at a fixed risk level. The key technique therein is to employ a sup-norm concentration of the empirical distribution (an improved version of the DKW inequality  \citep{massart1990tight}) to obtain a suitable VaR concentration result. The authors also provide a lower bound for the special case of the VaR at the level $0.75.$ A lower bound was then generalized to any level $\alpha$ in  \cite{david2016pure}. Notably, \cite{david2016pure} also proposes two PAC algorithms (namely Maximal Quantile and Doubled Maximal Quantile)  that have sample complexity within logarithmic factors of the lower bound. 

In~\cite{david2018pac}, the authors study a PAC problem with risk constraints. Therein, the objective is to find an arm with the highest mean reward (similar to the classic best arm identification problem), but only among those arms that satisfy a risk constraint. The risk measure is defined in terms of the $\vaR_{\alpha}(\cdot)$ of the arms being no smaller than a given $\beta.$ The authors propose a confidence interval-based algorithm and prove an upper bound on its sample complexity for sub-Gaussian arms' distribution. The upper bound has a similar form to the guarantee available for the risk-neutral PAC bandits problem \citep{even2006action}. They also show a lower bound on the sample complexity that matches the upper bound within logarithmic factors, using specific problem instances inspired by similar approaches in the risk-neutral setting \citep{mannor2004sample}.  An improvement to the algorithm in~\cite{david2018pac} based on the LUCB algorithm~\citep{Kalyanakrishnan2012}  was proposed by~\cite{hou2022almost} recently. 
\section{Future challenges}
In this survey, we reviewed the state-of-the-art in risk-aware MAB problems.  This is likely just the tip of an enormous iceberg, and that there are many fertile avenues for further research. 

The concentration bounds presented herein based on relating the risk measure to Wasserstein distance are likely to be suboptimal in terms of the constants. One can aim to tighten these bounds, which will have impact and utility beyond the study of MABs. 

One can explore methods other than the empirical estimators for various risk measures. For example, the potential of using importance sampling for estimating tail-based risk measures such as the CVaR has hitherto not been explored adequately. 

If improved concentration bounds can be derived, we can then apply them to obtain improved, problem-dependent bounds in both in regret minimization and best arm identification settings. These will also ascertain the asymptotic optimality of various algorithms by comparing the achievable bounds to the information-theoretic limits. 

From   Section~\ref{sec:bai}, we notice that there is generally much less work on pure exploration with risk, and only VaR and CVaR have been explored.   A unification of the algorithm design and accompanying guarantees for various risk measures would be welcome.

Finally, from a philosophical perspective, it is natural to wonder which risk measure to use when faced with a certain application scenario, such as portfolio optimization or clinical trials. A deeper, quantitative understanding of various risk measures and the ensuing implications on the scenario at hand requires more inter-disciplinary research.  Even if one is content with restricting oneself to say the mean-variance or CVaR, there still remains the issue of choosing an appropriate risk tolerance parameter $\gamma$ or the quantile level $\alpha$. 
\subsection*{Acknowledgements}
VT is supported by a Singapore National Research Foundation
(NRF) Fellowship (A-0005077-00-00) and Singapore Ministry
of Education AcRF Tier 1 grants (A-0009042-00-00, A-8000189-00-00,
and A-8000196-00-00).
\bibliographystyle{plainnat}
\bibliography{riskbib}
\end{document}